\def\year{2020}\relax
\documentclass[letterpaper]{article} 
\usepackage{aaai20}  
\usepackage{times}  
\usepackage{helvet} 
\usepackage{courier}  
\usepackage[hyphens]{url}  
\usepackage{graphicx} 
\usepackage{graphicx}  
\usepackage{amsfonts}
\usepackage{subfigure}
\usepackage{multirow}

\title{Multi-Component Graph Convolutional Collaborative Filtering}
\author{
Xiao Wang\textsuperscript{\rm 1},
Ruijia Wang\textsuperscript{\rm 1},
Chuan Shi\textsuperscript{\rm 1}\thanks{Corresponding author: Chuan Shi (shichuan@bupt.edu.cn)},
Guojie Song\textsuperscript{\rm 2},
Qingyong Li\textsuperscript{\rm 3}\\
\textsuperscript{\rm 1}Beijing University of Posts and Telecommunications,
\textsuperscript{\rm 2}Peking University,
\textsuperscript{\rm 3}Beijing Jiaotong University\\
\{xiaowang, wangruijia, shichuan\}@bupt.edu.cn, gjsong@pku.edu.cn, qingyongli@gmail.com
}

\begin{document}
\maketitle

\begin{abstract}
The interactions of users and items in recommender system could be naturally modeled as a user-item bipartite graph. In recent years, we have witnessed an emerging research effort in exploring user-item graph for collaborative filtering methods. Nevertheless, the formation of user-item interactions typically arises from highly complex latent purchasing motivations, such as high cost performance or eye-catching appearance, which are indistinguishably represented by the edges. The existing approaches still remain the differences between various purchasing motivations unexplored, rendering the inability to capture fine-grained user preference. Therefore, in this paper we propose a novel Multi-Component graph convolutional Collaborative Filtering (MCCF) approach to distinguish the latent purchasing motivations underneath the observed explicit user-item interactions. Specifically, there are two elaborately designed modules, decomposer and combiner, inside MCCF. The former first decomposes the edges in user-item graph to identify the latent components that may cause the purchasing relationship; the latter then recombines these latent components automatically to obtain unified embeddings for prediction. Furthermore, the sparse regularizer and weighted random sample strategy are utilized to alleviate the overfitting problem and accelerate the optimization. Empirical results on three real datasets and a synthetic dataset not only show the significant performance gains of MCCF, but also well demonstrate the necessity of considering multiple components.
\end{abstract}

\begin{figure}[t]
\centering
\includegraphics[width=.95\columnwidth]{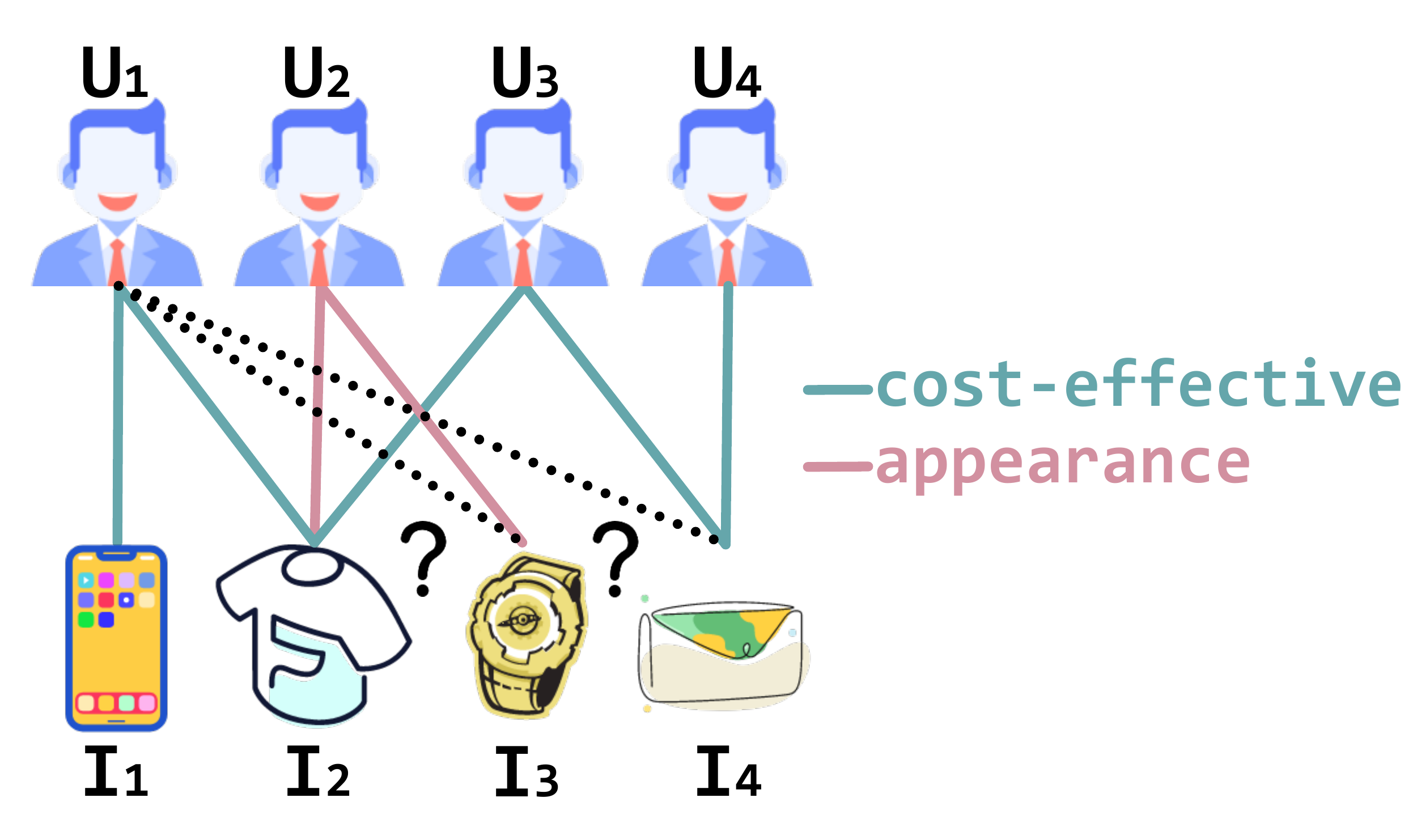} 
\caption{A toy example of purchasing relationships records with different purchasing motivations.}
\label{fig1}
\end{figure}
\section{Introduction}
Currently the overloaded online information overwhelms users. In order to tackle the information overload problem, Recommender Systems (RS) are widely employed to guide users in a personalized way of discovering products or services they might be interested in from a large number of possible alternatives. Due to its importance in practice, recommender systems have been attracting remarkable attention in both industry and academic research community \cite{berg2017graph}.

For many modern recommender systems, a de facto solution is the Collaborative Filtering (CF) technique, whose basic assumption is that people who share similar purchase in the past tend to have similar choice in the future \cite{koren2009matrix}. Essentially, the user-item interactions can be naturally modeled as a graph \cite{kalofolias2014matrix}, as exemplified by heterogeneous graph based recommendation \cite{shi2016integrating}. Therefore, graph convolutional networks \cite{defferrard2016convolutional,kipf2016semi}, which have demonstrated their remarkable ability in graph representation learning, are introduced to recommender systems and achieve promising performance \cite{berg2017graph,zheng2018spectral}. Typically, stacked graph convolutional layers are used on user-item graph to aggregate user and item features. In this way, these convolution operations can spread the information and fully take advantage of high-order relationship, thereby effectively alleviating the data sparsity problem in collaborative filtering.

Despite their enormous success, they all presume that user purchases items with uniform motivation, and ignore the fact that the formation of a real recommender system typically follows a complex and heterogeneous process, driven by the interactions of multiple \emph{latent components}. That is to say, although user-item interactions are all uniformly represented by the edges in user-item bipartite graph, there can be many different purchasing motivations for users to purchase items. For instance, different users may have different purchasing motivations, e.g., some prefer high cost performance, while some like eye-catching appearance. Basically, the user-item interaction system is not dominated by only one latent component (motivation), so treating all these latent components indistinguishably will inevitably lose some fine-grained valuable information. Considering the differences between purchasing motivations can capture more complex interaction characteristic and comprehensively reflect user preference, providing more accuracy recommendation clue.

Figure \ref{fig1} shows a toy example. The purchasing motivation of user $U_1$, $U_3$ and $U_4$ is high cost performance, while user $U_2$'s is eye-catching appearance. If ignoring latent components, it is not clear that user $U_1$ will purchase item $I_3$ or $I_4$. However, if we consider latent components, we may find item $I_4$ is a better recommendation to user $U_1$ because item $I_4$ has been purchased by value-oriented users, which is more in line with the user $U_1$ 's purchasing motivation. Hence failing to recognize the latent components underneath interactions may limit the recommendation performance. As a consequence, it is highly desired to design a new type of multi-component learner which can describe the fine-grained user preference.

Although it is promising to comprehensively utilize multiple latent components, it still faces the following two challenges. (1) \emph{How to identify multiple components in a user-item graph}? The user-item interaction system is highly complex, while the multiple components usually can not be directly observed. We need to effectively discover the corresponding latent component causing a specific purchasing relationship. Moreover, the extracted latent components should reflect different fine-grained user preference and embody rich semantics. (2) \emph{How to reorganize the multiple latent components to learn the user (item) embedding}? Even if we can extract multiple components, however, different users may be diverse in the selection of components. Therefore, effectively fusing these components is still a severe challenge.

In this paper, we propose a novel \textbf{M}ulti-\textbf{C}omponent Graph Convolutional \textbf{C}ollaborative \textbf{F}iltering (\textbf{MCCF}) approach, an end-to-end deep model that considers the diversity and heterogeneity of latent components in a uniform framework. Particularly, the key ingredient of MCCF consists of two modules, decomposer and combiner. Given a user-item interaction (edge), decomposer is to identify the latent components by decomposing the edge into multiple latent spaces with a node-level attention. The combiner is then to automatically determine the importance of these latent components and combine them to obtain the unified user (item) embeddings. Moreover, to cope with the overparameterization and overfitting problem, a sparse regularizer is utilized; to deal with noisy pairwise labels and accelerate the optimization, a weighted random sample strategy based on ratings is utilized meanwhile. 

We make the following contributions in this paper:
\begin{quote}
\begin{itemize}
\item We first study the problem of exploring multiple latent purchasing components in recommender system to capture more fine-grained user preference, given only explicit user-item interaction graph.
\item We propose MCCF, a novel collaborative filtering approach based on graph neural networks, to decompose and recombine the latent components underneath the edges of user-item graph. Moreover, the sparse regularizer and weighted random sample strategy are utilized to handle overparameterization and accelerate optimization, respectively.
\item We conduct extensive experiments on three real datasets and one synthetic dataset, which show the state-of-the-art performance of MCCF and the necessity of multiple components.
\end{itemize}
\end{quote}

\section{Related Work}
Our work bridges two very active areas of research: collaborative filtering and graph neural networks. Therefore, we mainly review the most related papers in the two areas. 

\begin{figure*}[t]
\centering
\includegraphics[width=.95\textwidth]{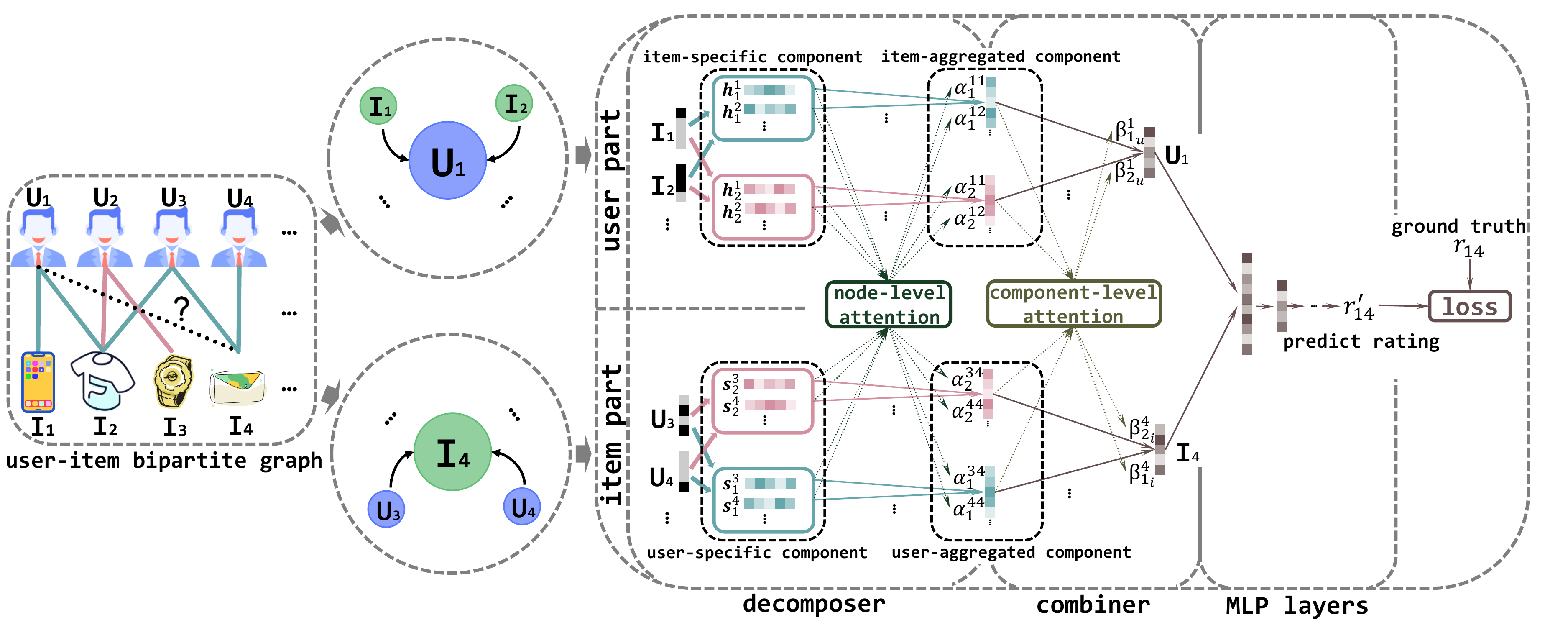} 
\caption{The framework of Multi-Component Graph Convolutional Collaborative Filtering (MCCF). It takes in the user-item bipartite graph and predicts user-item interaction ratings. This example assumes that there are two latent components, and predicts the rating that user $U_1$ would give to item $I_4$.}
\label{fig2}
\end{figure*}
\subsection{Collaborative Filtering}
Most popular collaborative filtering algorithms are based on matrix factorization (MF). Basically, they assume the rating matrix can be approximated by two low rank matrices. PMF \cite{mnih2008probabilistic} optimizes the maximum likelihood through minimizing the mean squared error between the observed entries and the reconstructed ratings. BiasedMF \cite{koren2009matrix} improves PMF by incorporating a user and item specific bias, as well as a global bias. Local low rank matrix approximation \cite{lee2013local} reconstructs rating matrix entries using different combinations of low rank approximations. Recently, with the surge of deep learning technique, neural MF models appear, such as AutoRec \cite{sedhain2015autorec} where the user's (item's) partially observed rating vector is projected onto a latent space through an encoder layer and reconstructs using a decoder layer; CF-NADE \cite{zheng2016neural} drops out part of input space at random in every iteration, which can be seen as a denoising auto-encoder. Another line is to treat the user-item interactions as graph to infer user preference. Early efforts attempt to leverage the compatibility of graph to fuse additional information, such as social information \cite{zhao2015connecting,zhao2014we} and heterogeneous information \cite{shi2016integrating}. Recently, deep network models are also employed to extract refined features from the user-item graph. GC-MC \cite{berg2017graph} uses two multi-link graph convolution layers to aggregate user features and item features. SpectralCF \cite{zheng2018spectral} proposes a spectral convolution operation to discover all possible connectivity between users and items. 
All these methods only treat the edge as a bridge to connect users and items without distinguishing multiple latent components. 

\subsection{Graph Neural Networks}
Graph neural networks (GNNs) \cite{gori2005new,scarselli2008graph}, especially graph convolutional networks \cite{bruna2013spectral,henaff2015deep}, have been attracting considerable attention lately, because of their remarkable success in various graph analysis tasks. The early attempts \cite{bruna2013spectral,henaff2015deep} to derive a graph convolutional layer are based on graph spectral theory, graph Fourier transformation \cite{shuman2012emerging} in particular. Then polynomial spectral filters are used to greatly reduce the computational cost \cite{defferrard2016convolutional}, and the usage of a linear filter makes further simplification \cite{kipf2016semi}. Along with spectral graph convolution, directly performing graph convolution in the spatial domain is also investigated \cite{duvenaud2015convolutional,atwood2016diffusion}. Later the attention mechanism \cite{bahdanau2014neural} is employed to adaptively specify weights to the neighbors of a node when performing spatial convolution \cite{velivckovic2017graph}. And heterogeneous graph attention network \cite{wang2019heterogeneous} is also used to fully consider the different importance of node and meta-path in the convolution process.

DisenGCN \cite{ma2019disentangled} is proposed to learn disentangled node representations, which employs a novel neighborhood routing mechanism to find the factor that may have caused the edge from a given node to one of its neighbors. However, DisenGCN is a homogeneous graph representation learning method, which does not distinguish the different importance among latent components meanwhile.

\section{Methodology}
\subsection{Overview}
Figure \ref{fig2} shows the overall framework of Multi-Component graph convolutional Collaborative Filtering (MCCF). As we can see, the model takes the user-item bipartite graph as input and predicts user-item interaction ratings. Specifically, the user part will aggregate the purchased item features to learn user embedding. During the item feature aggregation, we consider multiple latent components underneath the explicit user-item interactions through the following two modules: (1) a decomposer with node-level attention that identifies the latent components from the item feature; (2) a combiner with component-level attention that recombines the above latent components to obtain unified user embedding. When aggregating the user feature to learn the item embedding, we have the similar procedure as above. Finally, the MLP layers are applied to the learned user and item embeddings to output the rating. 

\subsection{Decomposer}
Here we give a formal definition of the user-item bipartite graph, followed by the description on the whole model from the user part, since the user and item parts are dual.

\subsubsection{User-Item Bipartite Graph}
In a recommendation scenario, we can typically model the historical user-item ratings as a user-item bipartite graph $\mathcal{G} = \{\mathcal{U}, \mathcal{I}, \mathcal{R}, \mathcal{E}\}$, where $\mathcal{U}$ and $\mathcal{I}$ are the sets of $N_u$ users and $N_i$ items; the rating set $\mathcal{R}$ may contain several ordinal rating levels $\{1, \cdots, R\}$. For each edge $e = (u, i, r) \in \mathcal{E}$, it represents that there is an observed rating value $r$ from user $u$ to item $i$. Generally, users have the feature matrix $\mathbf{U} = [\mathbf{u}_1, \mathbf{u}_2, \cdots, \mathbf{u}_{N_u}] \in \mathbb{R}^{L_u \times N_u}$, where $L_u$ is the dimension of user feature; items have feature matrix $\mathbf{P} = [\mathbf{p}_1, \mathbf{p}_2, \cdots, \mathbf{p}_{N_i}] \in \mathbb{R}^{L_i \times N_i}$, where $L_i$ is the dimension of item feature. 

\subsubsection{Multi-Component Extraction}
We assume that the user-item bipartite graph $\mathcal{G}$ is driven by $M$ latent components. The $m$-th component captures the $m$-th latent purchasing motivation in the user-item interactions. Therefore, we first design $M$ component-specific transformation matrices for the user and item respectively to extract different features that correspond to particular components, i.e. $\mathbf{W} = \{\mathbf{W}_1, \mathbf{W}_2, \cdots, \mathbf{W}_M\}$ and $\mathbf{Q} = \{\mathbf{Q}_1, \mathbf{Q}_2, \cdots, \mathbf{Q}_M\}$. For the item $i$, its $m$-th item-specific component $\mathbf{h}_m^i$ can be extracted as:
\begin{equation}
\mathbf{h}_m^i = \mathbf{Q}_m \mathbf{p}_i.
\end{equation}
Similarly, for user $u$, its $m$-th user-specific component $\mathbf{s}_m^u$ can be extracted as:
\begin{equation}
\mathbf{s}_m^u = \mathbf{W}_m \mathbf{u}_u.
\end{equation}

\subsubsection{Node-Level Attention}
In the following, we focus on user $u$ and his purchased item set $\mathcal{P}_u$. For user $u$, there are $M$ user-specific components $\{\mathbf{s}_m^u\}_{m=1}^M$. For the item $i\in \mathcal{P}_u$, it also has $M$ item-specific components $\{\mathbf{h}_m^i\}_{m=1}^M$. The key insight here is that user $u$ does not need aggregate all the purchased items to describe $m$-th component. Therefore, we propose node-level attention mechanism to infer the items that are actually purchased by user $u$ due to $m$-th component.

Specifically, the possibility of user $u$ purchasing item $i$ based on the $m$-th component can be formulated as follows:
\begin{equation}
e_m^{ui} = att_{node}(\mathbf{s}_m^u, \mathbf{h}_m^i; m),
\label{eq3}
\end{equation}
where $att_{node}$ denotes the deep neural network which performs the node-level attention. Eq. (\ref{eq3}) shows that based on $m$-th component, the possibility of user $u$ purchasing item $i$ depends on their own features with this component. After obtaining the possibility $e_m^{ui}$, we normalize it to get the weight coefficient $\alpha_m^{ui}$ via softmax function:
\begin{equation}
\alpha_m^{ui} = softmax\left(e_m^{ui}\right) = \frac{\exp \left(\sigma\left(\mathbf{a}_m^{\mathrm{T}} \cdot\left[\mathbf{s}_m^u \| \mathbf{h}_m^i\right]\right)\right)}{\sum_{i \in \mathcal{P}_u} \exp \left(\sigma\left(\mathbf{a}_m^{\mathrm{T}} \cdot\left[\mathbf{s}_m^u \| \mathbf{h}_m^i\right]\right)\right)},
\end{equation}
where $\sigma$ denotes the activation function, $\|$ denotes the concatenate operation and $\mathbf{a}_m$ is the node-level attention vector for $m$-th component. 

Finally, for all the items in $\mathcal{P}_u$, by aggregating all their $m$-th item-specific components, we can learn the $m$-th item-aggregated component $\mathbf{z}_m^u$ for user $u$ as follows:
\begin{equation}
\mathbf{z}_m^u = \sigma\left(\sum_{i \in \mathcal{P}_u} \alpha_m^{ui} \cdot \mathbf{h}_m^i\right).
\end{equation}
Now, each user $u$ will have $M$ item-aggregated components $\{\mathbf{z}_m^u\}_{m=1}^M$. Please note that, every item-aggregated component of user $u$ is aggregated by the features of his purchased items under this component, thus it is semantic-specific and able to capture the corresponding purchasing motivation represented by the component. Next, we will introduce how to combine $\{\mathbf{z}_m^u\}_{m=1}^M$ to learn the final user embedding.

\subsection{Combiner}
It is well recognized that a user's purchasing behavior is usually driven by one or some motivations, which can be reflected by the learned item-aggregated components. Therefore, different components should have different contributions to learn the user embedding, motivating us to propose a combiner with component-level attention mechanism to automatically learn the importance of different item-aggregated components. 
\subsubsection{Component-Level Attention}
Taking $M$ item-aggregated components of user $u$ as input, we aim to learn the weights of each item-aggregated component $\left(\beta_1^u, \beta_2^u, \cdots, \beta_M^u\right)$ as follows:
\begin{equation}
\left(\beta_1^u, \beta_2^u, \cdots, \beta_M^u\right) = att_{com}\left(\mathbf{z}_1^u, \mathbf{z}_2^u, \cdots, \mathbf{z}_M^u\right),
\end{equation}
where $att_{com}$ denotes the deep neural network which performs the component-level attention. It shows that the component-level attention can differentiate the importance of item-aggregated components.

Considering that the importance of the $m$-th item-aggregated component $\beta_m^u$ should also depend on the user $u$, we first concatenate $\mathbf{z}_m^u$ and $\mathbf{s}_m^u$, and learn their unified embedding as follows:
\begin{equation}
\mathbf{d}_m^u = \sigma\left(\mathbf{C}_m\cdot\left[\mathbf{z}_m^u \| \mathbf{s}_m^u\right] + \mathbf{b}_m\right),
\end{equation}
where $\mathbf{C}_m$ is the weight matrix and $\mathbf{b}_m$ is the bias vector. Then with a component-level attention vector $\mathbf{q}$, the importance of the $m$-th item-aggregated component, denoted as $w_m$, is shown as follows:
\begin{equation}
w_m = \sigma \left(\mathbf{q}^{\mathrm{T}} \cdot \mathbf{d}_m^u + b\right),
\end{equation}
where $b$ is bias. Note that parameters $\mathbf{q}$ and $b$ are shared for all users and item-aggregated components, which is because there are some similar decision patterns in human nature while purchasing the items. We then normalize $w_m$ via softmax function to obtain the weight of $m$-th item-aggregated component $\beta_m^u$ as follows:
\begin{equation}
\beta_m^u = \frac{\exp \left(w_m\right)}{\sum_{k=1}^M \exp \left(w_k\right)}.   
\end{equation}
 Obviously, the higher $\beta_m^u$, the purchasing relationship more likely to be caused by the $m$-th purchasing component. 
 
 With these learned weights, we can fuse these item-aggregated components to obtain the final embedding $\mathbf{z}_u$ of user $u$ as follows:
\begin{equation}
\mathbf{z}_u = \sum_{m=1}^{M} \beta_m^u \cdot \mathbf{z}_m^u.
\end{equation}

\textbf{Remark}: We elaborately describe the user representation learning process here. Because the item representation learning is a dual process, we omit it for brevity.

\subsection{Rating Prediction}
Once obtaining the final embeddings of user $u$ and item $i$ from the user and item part separately (i.e., $\mathbf{z}_u$ and $\mathbf{v}_i$), we concatenate them and make it pass through MLP to predict the rating $r_{ui}^{\prime}$ from $u$ to $i$ as:
\begin{eqnarray}
\mathbf{g}_1 &=& \left[\mathbf{z}_u \| \mathbf{v}_i\right], \\ 
\mathbf{g}_2 &=& \sigma\left(\mathbf{W}_2 \cdot \mathbf{g}_1 + \mathbf{b}_2\right), \\ 
&\cdots& \\ 
\mathbf{g}_{l-1} &=&\sigma\left(\mathbf{W}_l \cdot \mathbf{g}_{l-1}+\mathbf{b}_l\right), \\ 
r_{ui}^{\prime} &=& \mathbf{w}^T \cdot \mathbf{g}_{l-1},
\end{eqnarray}
where $l$ is the index of a hidden layer.

\subsection{Optimization}
\subsubsection{Objective Function}
Since the task is the rating prediction, the primary goal is to minimize the difference of predicted ratings and ground truth:
\begin{equation}
\mathcal{L}_r = \frac{1}{2|\mathcal{O}|} \sum_{(u, i) \in \mathcal{O}}\left(r_{ui}^{\prime}-r_{ui}\right)^{2},
\end{equation}
where $\mathcal{O}$ is the set of observed ratings, and $r_{ui}$ is the ground truth rating by the user $u$ on the item $i$.

Then it is worth noting that to alleviate overparametrization and overfitting, the multiple components should be properly regularized. Therefore, we employ the $L_0$ regularization \cite{louizos2017learning} to our objective function. By sparsifying the multi-component extraction matrices $\mathbf{W}$ and $\mathbf{Q}$, we can avoid unnecessary resources and alleviate overfitting, because irrelevant degrees of freedom are pruned away. The final objective function is as follows:
\begin{equation}
\min _{\Theta} \mathcal{L} = \mathcal{L}_r + \lambda\left\|\theta\right\|_0,
\end{equation}
where $\Theta$ denotes the model parameter set, $\theta = \{\mathbf{W}, \mathbf{Q}\}$ and $\lambda$ is a hyper-parameter to balance the rating loss and sparse regularization. 

\subsubsection{Sample Strategy}
It is well known that items with high ratings better reflect user preference, and we may not need to aggregate all the purchased items of users in practice. Therefore, we employ a weighted random sampling. The sampling can pay more attention on high-rating user-item pairs and accelerate model optimization meanwhile. 

Specifically, we calculate the average degree $N^u$ ($N^i$) of user (item) node in the bipartite graph, which is used as the user (item) threshold. When the number of neighbors exceeds the threshold, the sample strategy is applied, otherwise all neighbors are retained. The sampling process \cite{efraimidis2006weighted} is as follows:
\begin{eqnarray}
u &\sim& \mathcal{U}(0,1), \\
k &=& u^{\frac{1}{r}},
\end{eqnarray}
where $\mathcal{U}$ denotes the uniform distribution, $r$ denotes the rating and $k$ is the generated weighted random number. We generate a corresponding $k$ for every neighbor, then sort them in the descending order. Finally, we select top-$N^u$ ($N^i$) neighbors for the convolution operation.

\section{Experiments}
We perform experiments on three real datasets and a synthetic dataset to evaluate our proposed model and answer the following questions:
\begin{quote}
\begin{itemize}
\item \textbf{Q1}: How does MCCF perform as compared with state-of-the-art collaborative filtering methods?
\item \textbf{Q2}: Does the embedding learned from multiple components have stronger representation capability than the undecomposed? Could multiple components capture some latent semantics?
\item \textbf{Q3}: How do different hyper-parameters settings affect MCCF?
\end{itemize}
\end{quote}

\begin{table*}[!t]
\centering
\caption{Performance comparison of rating prediction. The smaller values, the better performance.}\smallskip
\resizebox{\textwidth}{!}{
\begin{tabular}{cc|cccccc|ccc}
\hline
\multicolumn{2}{c|}{\textbf{Models}} & \textbf{PMF} & \textbf{BiasMF} & \textbf{LLORMA-Local} & \textbf{I-AUTOREC} & \textbf{I-CF-NADE} & \textbf{GC-MC} & \textbf{MCCF}-$nd$ & \textbf{MCCF}-$cmp$ & \textbf{MCCF} \\ \hline
\multirow{2}{*}{\textbf{Yelp}} & \textbf{RMSE} & 0.3967 & 0.3902 & 0.3890 & 0.3817 & 0.3857 & 0.3850 & 0.3836 & \textbf{0.3806} & \textbf{0.3806} \\  
 & \textbf{MAE} & 0.1571 & 0.1616 & 0.1547 & 0.1201 & 0.1427 & 0.1354 & 0.1286 & \textbf{0.1029} & \textbf{0.1029} \\ \hline
\multirow{2}{*}{\textbf{Amazon}} & \textbf{RMSE} & 0.9339 & 0.9028 & 0.9019 & 0.9213 & 0.8987 & 0.8946 & 0.8942 & 0.8919 & \textbf{0.8876} \\  
 & \textbf{MAE} & 0.7113 & 0.6759 & 0.6725 & 0.7064 & 0.6565 & 0.6619 & 0.6595 & 0.6483 & \textbf{0.6428} \\ \hline
\multirow{2}{*}{\textbf{Movielens}} & \textbf{RMSE} & 0.9638 & 0.9257 & 0.9313 & 0.9435 & 0.9229 & 0.9145 & 0.9203 & 0.9142 & \textbf{0.9070} \\  
 & \textbf{MAE} & 0.7559 & 0.7258 & 0.7286 & 0.7370 & 0.7168 & 0.7165 & 0.7160 & 0.7081 & \textbf{0.7050} \\ \hline
\end{tabular}
}
\label{tab1}
\end{table*}
\subsection{Experimental Settings}
\subsubsection{Datasets}
We conduct experiments on three real datasets: MovieLens, Amazon and Yelp, which are publicly accessible and vary in terms of domain, size and sparsity.
\begin{quote}
\begin{itemize}
\item \textbf{MovieLens-100k}: A widely adopted benchmark dataset in movie recommendation, which contains 100,000 ratings from 943 users to 1, 682 movies.
\item \textbf{Amazon}: A widely used product recommendation dataset, which contains 65, 170 ratings from 1, 000 users to 1, 000 items.
\item \textbf{Yelp}: A local business recommendation dataset, which contains 30, 838 ratings from 1, 286 users to 2, 614 items.
\end{itemize}
\end{quote}
For each dataset, we randomly select 80\% of historical ratings as training set, and treat the remaining as test set. 

\subsubsection{Baselines}
We compare MCCF with several state-of-the-arts, including matrix factorization methods: PMF \cite{mnih2008probabilistic}, BiasMF \cite{koren2009matrix} and LLORMA-Local \cite{lee2013local}; auto-encoders based methods: AUTOREC \cite{sedhain2015autorec} and CF-NADE \cite{zheng2016neural}; graph convolutional networks based collaborative filtering model: GC-MC \cite{berg2017graph}. Typically, we use I-AUTOREC and I-CF-NADE to represent the item-based setting, which has better performance than the user-based. In addition, we also adopt two variants of MCCF (MCCF-$nd$ and MCCF-$cmp$) as baselines to analyze the role of hierarchical attention. Specifically, MCCF-$nd$ removes the node-level attention in decomposer, while MCCF-$cmp$ removes the component-level attention in combiner.

\subsubsection{Implementation}
We consider the feature matrix $\mathbf{X}$ as the adjacency matrix. And we vary the number of components $K$ in range $\{1, 2, 3, 4\}$ and the embedding dimension $d$ in range $\{8, 16, 32, 64, 128\}$. For neural network, we empirically employ two layers for all the neural parts and the activation function as ReLU. We randomly initialize the model parameters with a Gaussian distribution $\mathcal{N}(0, 0.1)$, then use the Adam as the optimizer. The batch size and learning rate are searched in $\{64, 128, 256, 512\}$ and $\{0.0005, 0.001, 0.002, 0.0025\}$, respectively. Meanwhile, the dropout is applied to our model except for multi-component extraction, and the dropout rate is tested in $\{0.1, 0.4, 0.5, 0.6\}$. The parameters for $L_0$ regularization are set according to literature \cite{louizos2017learning}. All the baselines are initialized as the corresponding papers, and in terms of neural network models we use the same embedding dimension for fair comparison. Then they are carefully tuned to achieve optimal performance. We adopt two widely-used evaluation protocols: \emph{Root Mean Squard Error} (RMSE) and \emph{Mean Absolute Error} (MAE) as evaluation metrics. We repeat five runs with random initialization for all models and report the average results.

\subsection{Performance Comparison (Q1)}
We first compare the recommendation performance of all methods. Table \ref{tab1} shows the overall rating prediction error. We have the following observations: (1) Our model MCCF consistently outperforms all the baselines, suggesting the effectiveness of MCCF on recommendation. (2) By comparing with MCCF, we find that the performances of MCCF-$nd$ and MCCF-$cmp$ have various degrees of degeneration except for MCCF-$cmp$ on Yelp, for reason that the number of latent components on Yelp is one according to the optimal experiment setting. These results are consistent with the two assumptions of MCCF, namely not all purchased items of one user contribute equally to the different latent components and not all latent components have the same importance to learning the final embeddings. This phenomenon also demonstrates the benefits of the hierarchical attention. (3) We observe that I-AUTOREC, I-CF-NADE and GC-MC generally outperform PMF, BiasMF and LLORMA-Local, suggesting the power of neural network models. Meanwhile among these baselines, GC-MC shows quite strong performance, which implies that the GNNs are powerful in representation learning for graph data.

\begin{figure}[!t]
\centering
\includegraphics[width=.45\columnwidth]{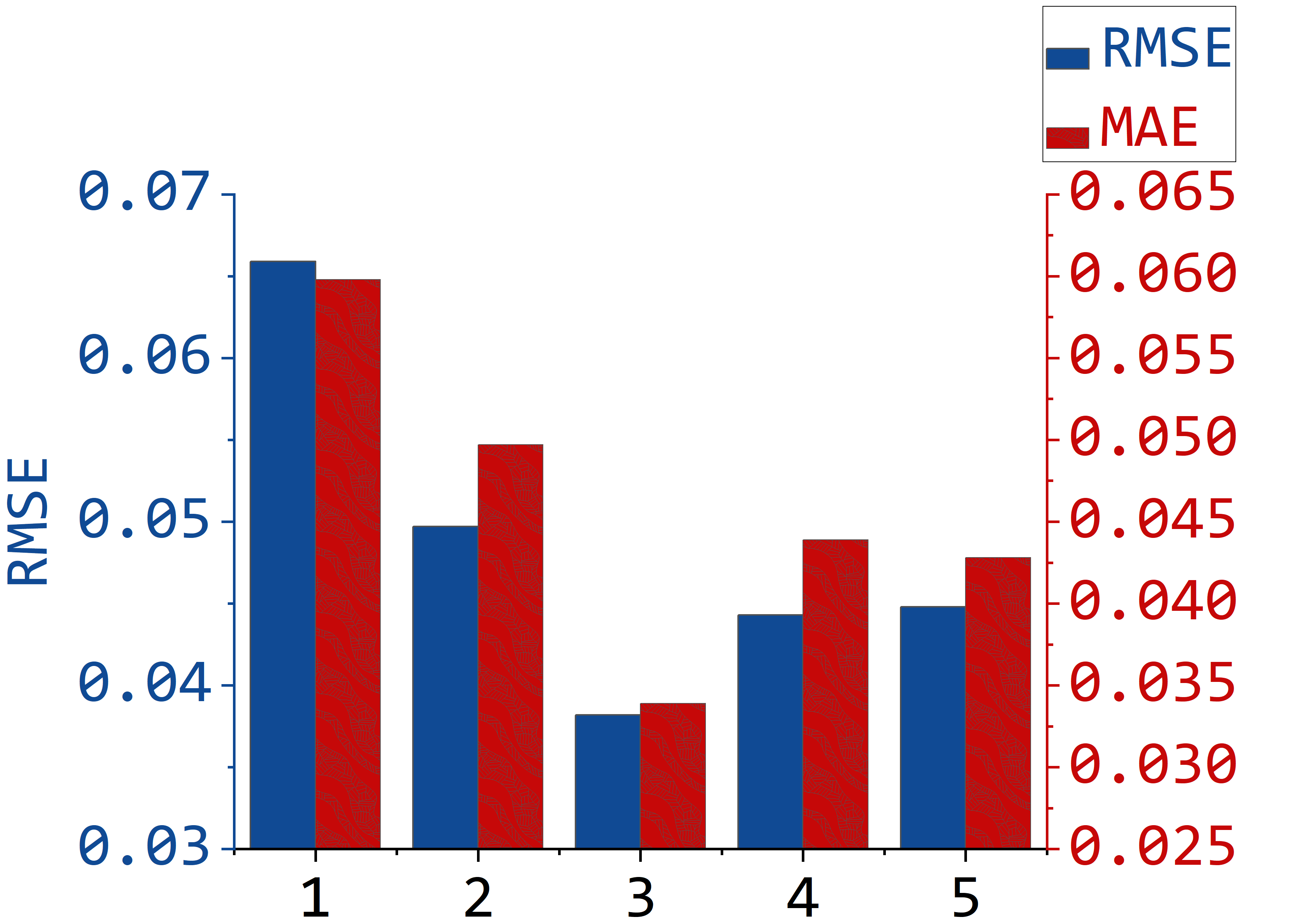} 
\caption{RMSE and MAE on synthetic user-item bipartite graph generated with three latent components.}
\label{fig3}
\end{figure}
\subsection{Effect of Multiple Components (Q2)}
We further generate a synthetic user-item bipartite graph to investigate the behavior of multiple components. Since the features of users and items in our model are equal to adjacency vectors, different latent components are distinguished by the sparsity of the adjacency vectors. Thus to generate a user-item graph with three latent components, we first generate three user-item subgraphs with different sparsity, each of which has 300 users and 100 items. User-item pairs in a user-item subgraph are connected if the absolute value of the number $p$ sampled from the Gaussian distribution exceeds the threshold 0.5. Corresponding to three subgraphs, we sample from three different Gaussian distributions, and their mean is 0, while variances are 0.5, 5 and 50 respectively. The 300 users among these subgraphs are shared, while items are disjointed. Then we generate the final synthetic graph with 300 users and 300 items by concatenating the adjacency matrices of the above three user-item subgraphs. And the ground-truth components of items are used as labels.
\begin{figure}[!t]
\centering
\subfigure[Node-level attention weights visualization for the 3$^{rd}$ epoch.]{
\includegraphics[width=.4\columnwidth]{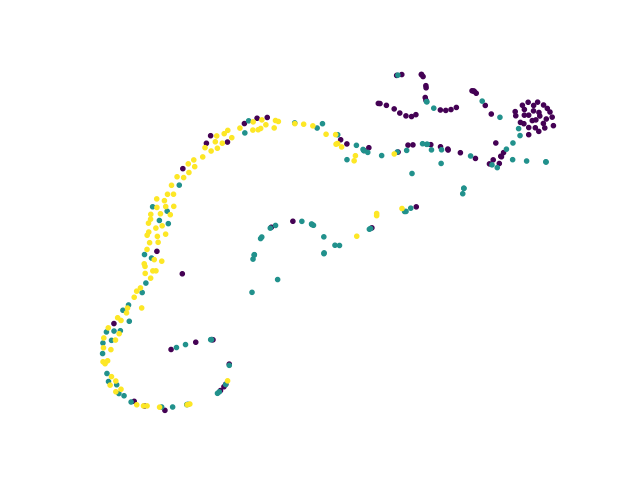}  
}     
\hspace{.4in}
\subfigure[Component-level attention weights visualization for the 3$^{rd}$ epoch.] { 
\includegraphics[width=.4\columnwidth]{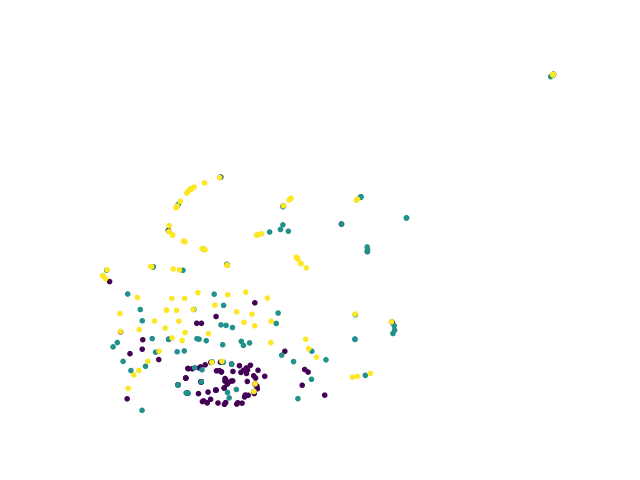}     
}    
\vfill
\subfigure[Node-level attention weights visualization for the 10$^{th}$ epoch.] { 
\includegraphics[width=.4\columnwidth]{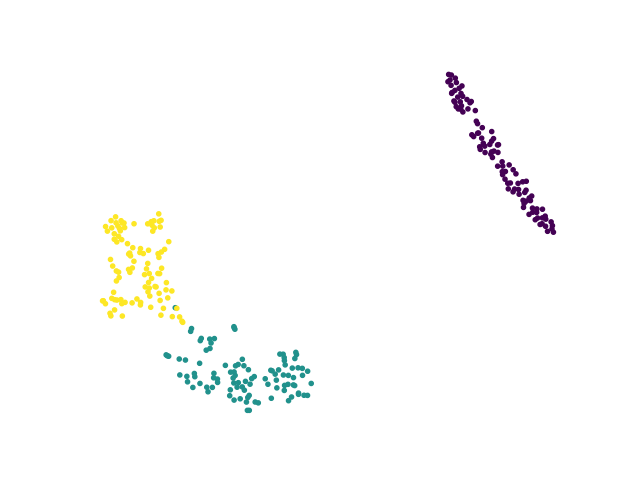}     
}   
\hspace{.4in}
\subfigure[Component-level attention weights visualization for the 10$^{th}$ epoch.] { 
\includegraphics[width=.4\columnwidth]{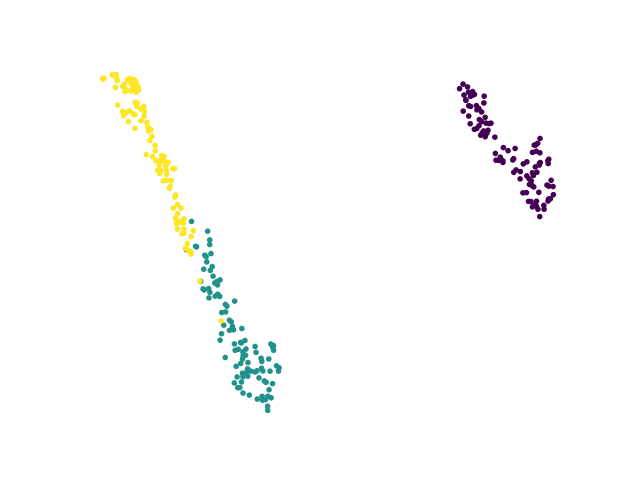}     
}  
\caption{Attention weights visualization of the synthetic dataset on different epochs. Each point indicates one attention weights distribution of item, and the color of a point indicates the class of the item.}
\label{fig4}
\end{figure}

\begin{figure*}[!t]
\centering
\subfigure[Yelp.]{
\includegraphics[width=.19\textwidth]{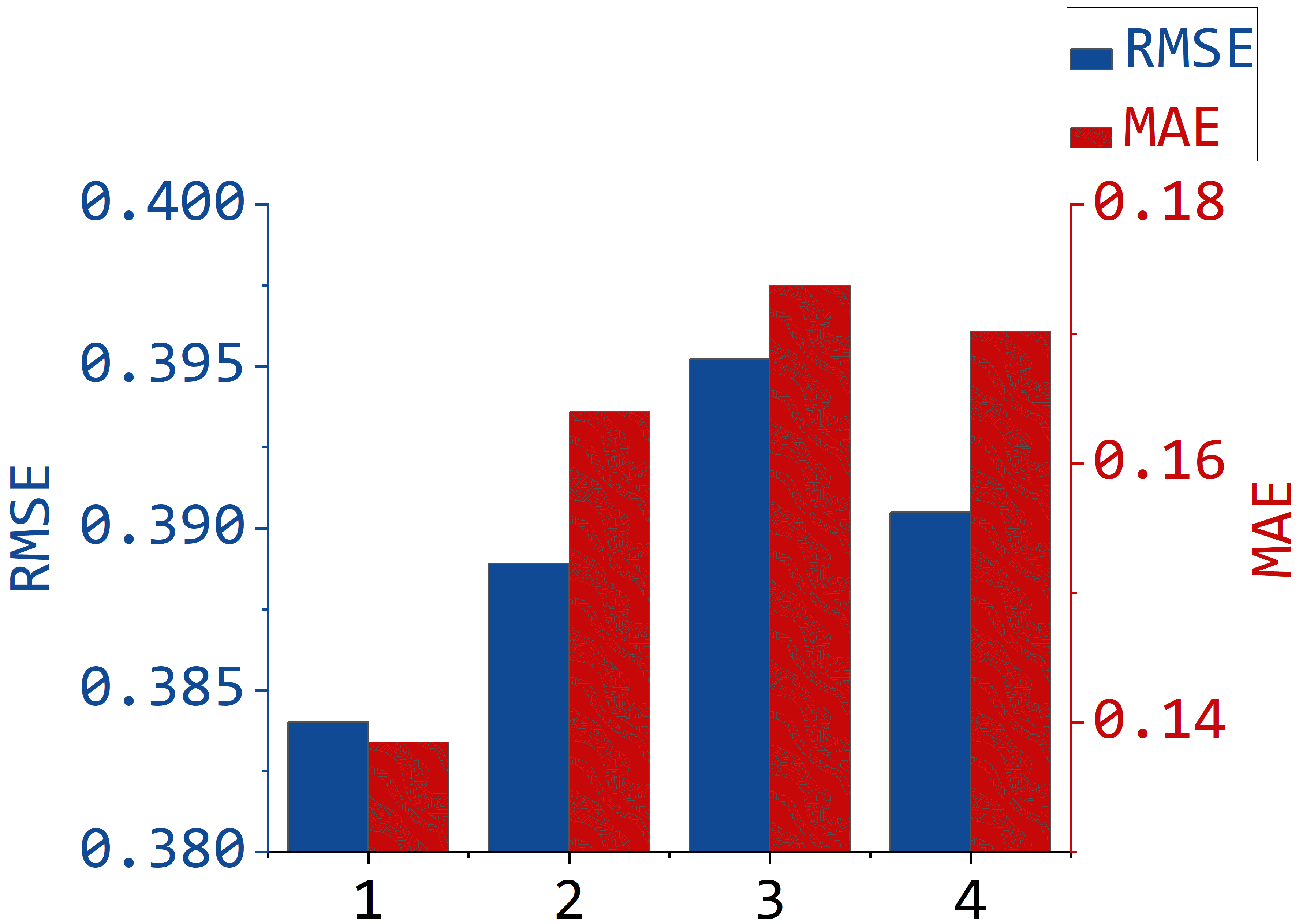}  
}     
\hspace{.9in}
\subfigure[Amazon.] { 
\includegraphics[width=.19\textwidth]{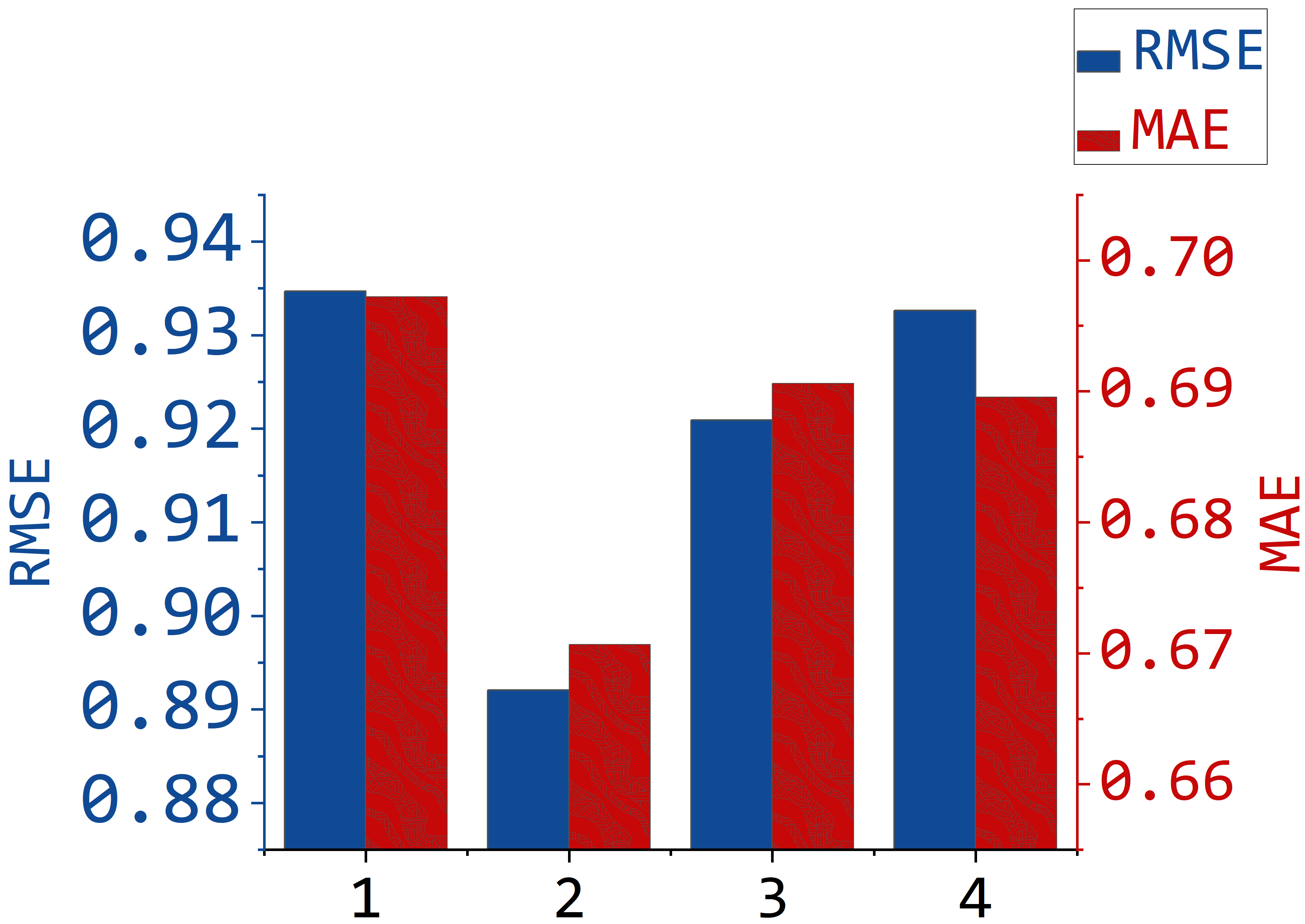}     
}    
\hspace{.9in}
\subfigure[MovieLens.] { 
\includegraphics[width=.19\textwidth]{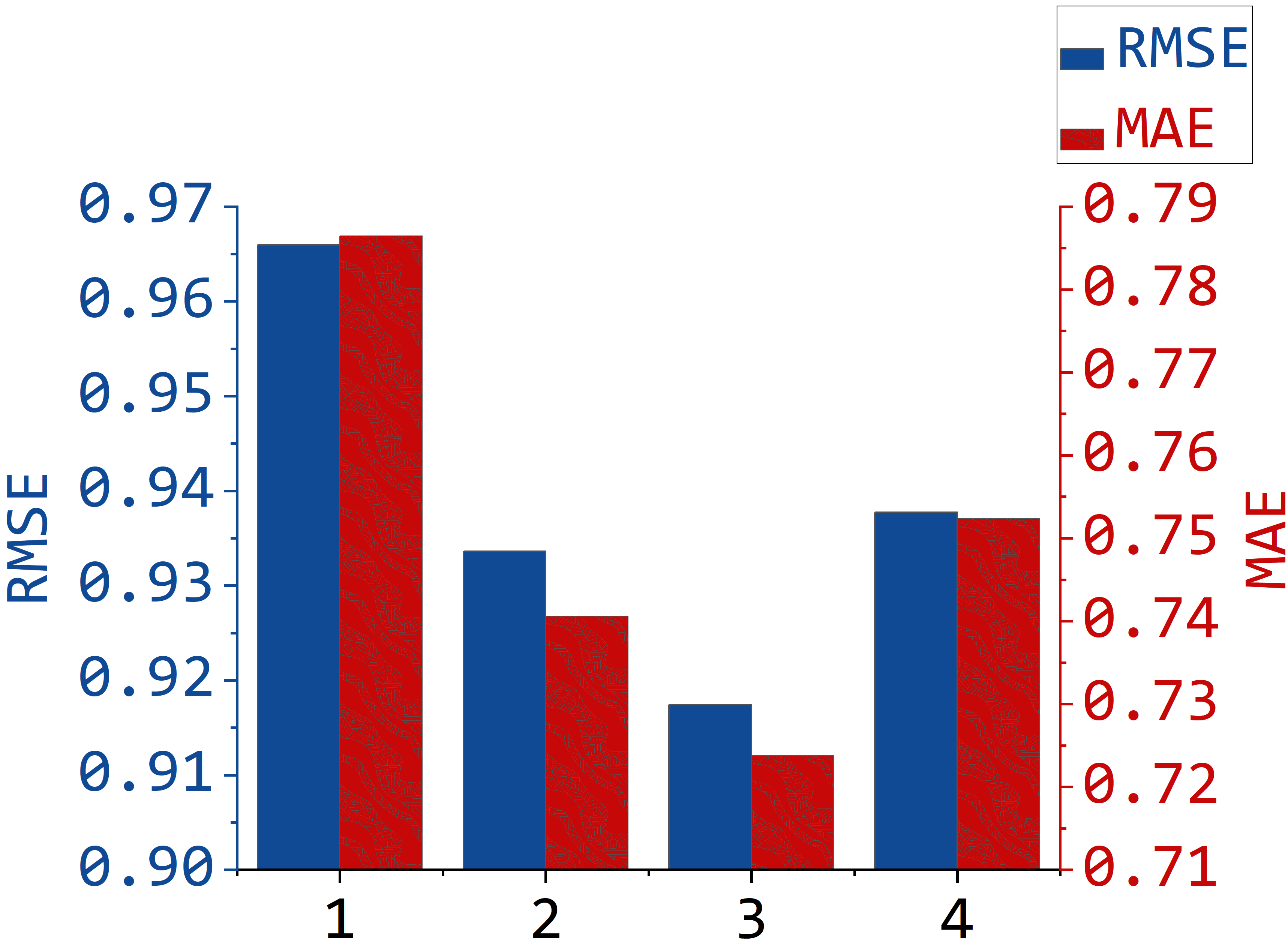}     
}  
\caption{Impact of latent components numbers on three real datasets.}
\label{fig5}
\end{figure*}

\begin{figure*}[!t]
\centering
\subfigure[Yelp.]{
\includegraphics[width=.19\textwidth]{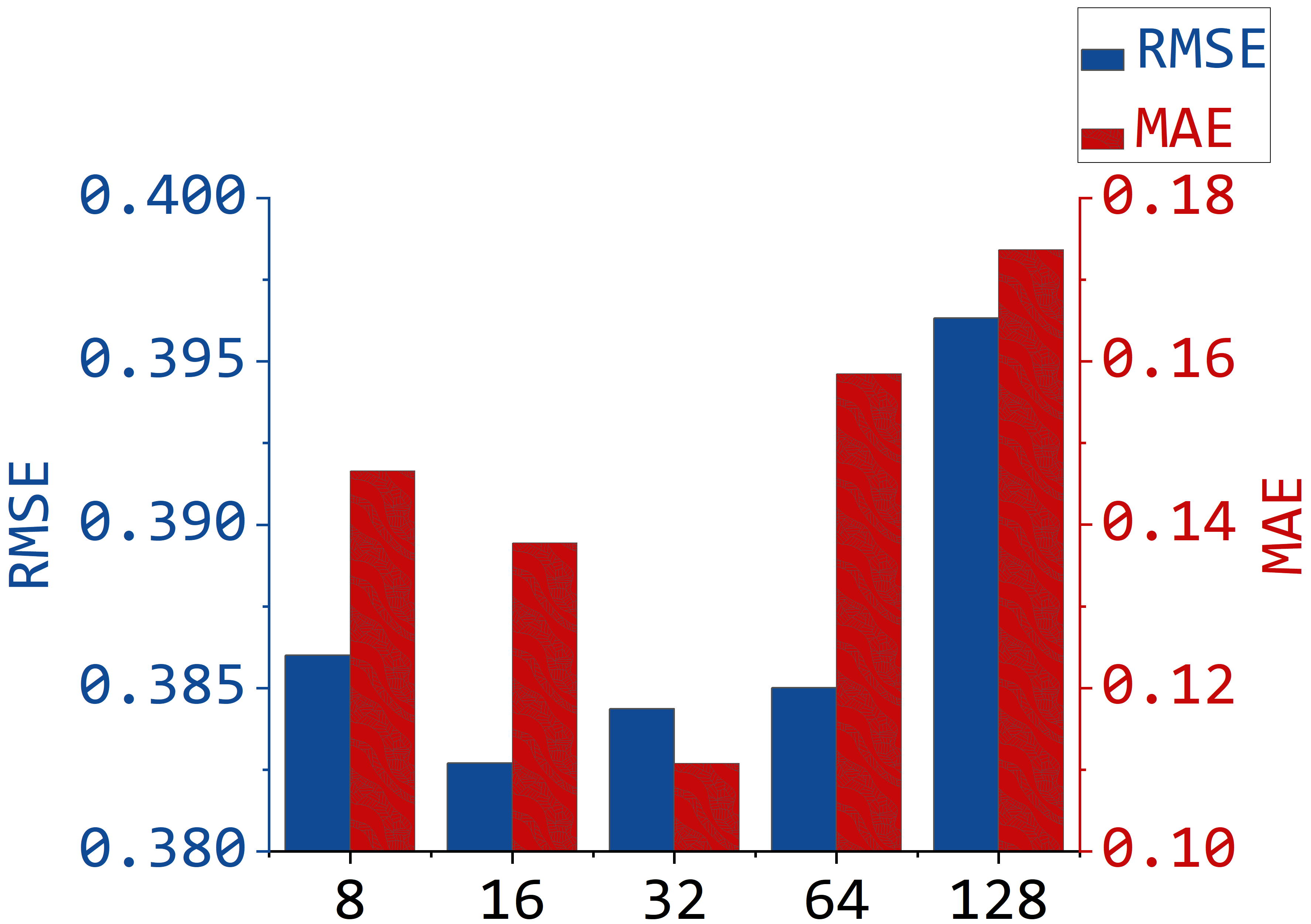}  
}     
\hspace{.9in}
\subfigure[Amazon.] { 
\includegraphics[width=.19\textwidth]{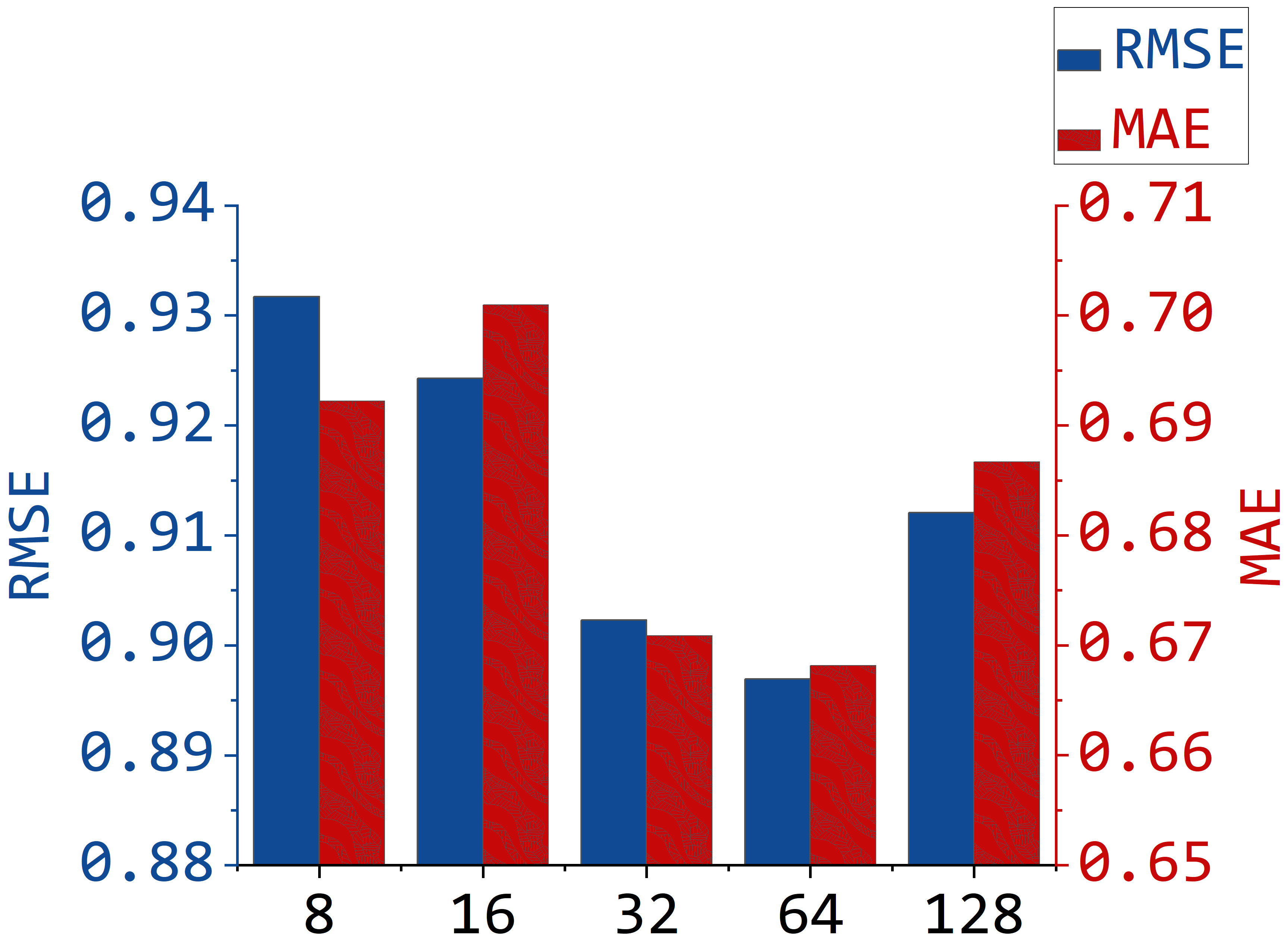}     
}    
\hspace{.9in}
\subfigure[MovieLens.] { 
\includegraphics[width=.19\textwidth]{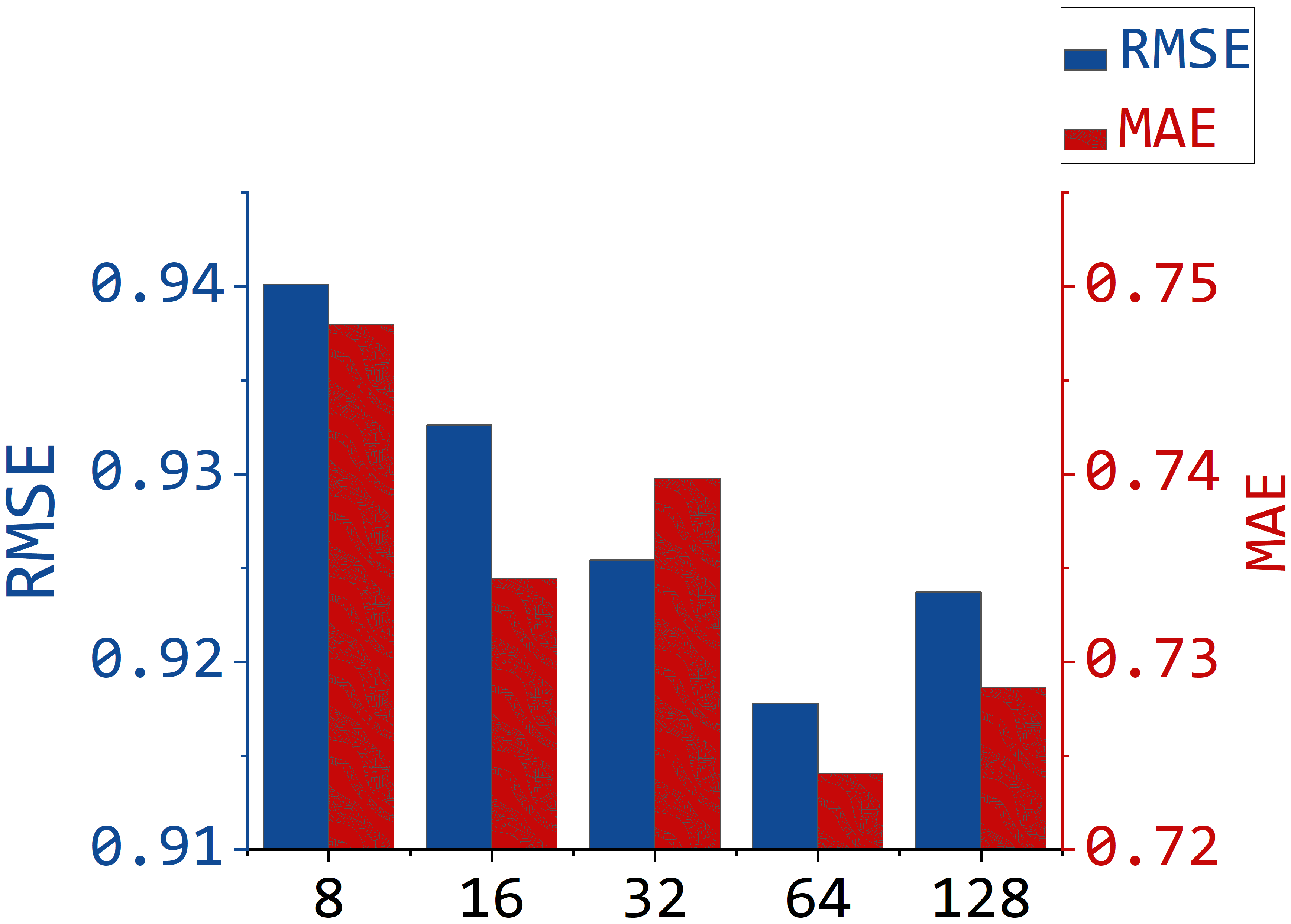}     
}  
\caption{Impact of embedding dimensions on three real datasets.}
\label{fig6}
\end{figure*}
\subsubsection{Consistency in the Number of Latent Components}
We vary the number of components $K$ from 1 to 5, while keeping the other parameters the same, and report the recommendation results in Figure \ref{fig3}. From the results, we find that as the number of latent components increases from 1 to 3, MCCF starts to achieve a greater improvement, indicating the importance of considering multiple components. In particular, when the number of components $K$ equals to 3, the best performance is achieved. This demonstrates the implicit semantic capturing capability of multiple components. However, when the number of components $K$ continues to grow, the performance is saturated and even drops.

\subsubsection{Attention Weights Visualization}
The hierarchical attention mechanism is also a key ingredient of MCCF. Therefore, to further verify the validity of the hierarchical attention and the implicit semantic capturing capability of multiple components, we apply the best performance setting on synthetic dataset, i.e. the number of components $K$ is 3, and use the attention weights learned by hierarchical attention as the input to the visualization tool t-SNE \cite{maaten2008visualizing}. For the node-level attention, we randomly select a user $u$ so that he is connected to all items, and visualize the node-level attention weights $\{\alpha_m^{ui}\}_{m=1}^M$ on item-specific components for every item $i$ in Figure \ref{fig4} (a) and (c). For the component-level attention, we visualize the component-level attention weights $\{\beta_m^i\}_{m=1}^M$ on user-aggregated components for every item $i$ in Figure \ref{fig4} (b) and (d). Specifically, according to the generation process of the user-item graph, we know that items are divided into three disjointed classes. Based on the MCCF premises, items of the same class should have similar weights distributions. A good visualization result is that the points of the same class are closer to each other. At the beginning of model training, the result is unsatisfactory since the points belonging to different classes are mixed with each other. And after a few epoches, we can observe clear clusters of different classes. This again validates the strong representation power of multiple components.

\subsection{Parameter Analysis of MCCF (Q3)}
As the number of components $K$ plays a pivotal role in MCCF, we investigate its impact on the performance, and then we analyze the influence of embedding dimension $d$.

\subsubsection{Impact of Latent Components Numbers}
To investigate whether MCCF can benefit from multiple components, we vary the number of components $K$ in the range of $\{1, 2, 3, 4\}$, while keeping the other parameters the same. Figure \ref{fig5} shows the experimental results in real datasets. We can see that the optimal number of components varies with the specific dataset. For Yelp, the user-item graph is extremely sparse, and most ratings are 1 or 2. Therefore, one component is enough to model latent semantics. As for Amazon and MovieLens, user-item graphs are much denser with an even distribution of ratings. At this point, the power of multiple components is more prominent. Increasing $K$ leads to performance improvement. After the best performance is reached, the improvement tends to become saturated and even drop as $K$ continues to grow, possibly due to overfitting problem.

\subsubsection{Impact of Embedding Dimensions}
The dimension of embeddings $d$ is also a key parameter to control the complexity and capacity of MCCF. Therefore, we evaluate how it affects the recommendation performance. In Figure \ref{fig6}, generally speaking, as we gradually increase embedding dimension $d$, the recommendation performance grows since a larger $d$ could enhance the representation capability. Nevertheless, when $d$ is larger than the optimal values, increasing $d$ will hurt the performance. Therefore, we employ the proper embedding dimension $d$ to balance the trade-off between performance and complexity.

\section{Conclusion}
We develop a novel recommender system model called Multi-Component graph convolutional Collaborative Filtering (MCCF). The idea is to explore the differences between purchasing motivations underneath the simple edges in user-item bipartite graph, where edges are decomposed and then recombined with hierarchical attention to encode the latent semantics based on the specific user-item pair. In contrast to standard holistic methods, multiple components significantly enrich the representation capability and reflect fine-grained user preference. Extensive experiments demonstrate that MCCF not only outperforms existing methods in terms of recommendation accuracy, but also captures the latent semantics in datasets. In future, we will work on the further improvement in optimization efficiency. In addition, we are interested in integrating auxiliary information to advance the performance, not limited to the structural information of the user-item graph. 

\section{Acknowledgments}
We thank anonymous reviewers of AAAI 2020 for their time and effort in reviewing this paper. This work is supported in part by the National Natural Science Foundation of China (No. 61702296, 61972442, 61772082, 61806020), the National Key Research and Development Program of China (2017YFB0803304), the Beijing Municipal Natural Science Foundation (4182043), 2019 and 2018 the CCF-Tencent Open Fund.

\bibliographystyle{aaai}
\bibliography{references}

\end{document}